\documentclass[10pt,twocolumn,letterpaper]{article}

\usepackage{iccv_21_template/iccv}
\usepackage{times}
\usepackage{epsfig}
\usepackage{graphicx}
\usepackage{amsmath}
\usepackage{amssymb}

\usepackage{authblk}

\usepackage[pagebackref=true,breaklinks=true,letterpaper=true,colorlinks,bookmarks=false]{hyperref}

\iccvfinalcopy 

\def\iccvPaperID{9080} 

\ificcvfinal\fi

\begin{document}

\title{Fully-Automatic Pipeline for Document Signature Analysis to Detect Money Laundering Activities }

\author[1]{Nikhil Woodruff}
\author[1,2]{Amir Enshaei}
\author[1]{Bashar Awwad Shiekh Hasan}
\affil[1]{Caspian Learning, Newcastle Upon Tyne, United Kingdom}
\affil[2]{Faculty of Medical Sciences, Newcastle University, Newcastle Upon Tyne, United Kingdom}

\maketitle
\ificcvfinal\thispagestyle{empty}\fi

\begin{abstract}
   Signatures present on corporate documents are often used in investigations of relationships between persons of interest, and prior research into the task of offline signature verification has evaluated a wide range of methods on standard signature datasets. However, such tasks often benefit from prior human supervision in the collection, adjustment and labelling of isolated signature images from which all real-world context has been removed. Signatures found in online document repositories such as the United Kingdom’s Companies House regularly contain high variation in location, size, quality and degrees of obfuscation under stamps. We propose an integrated pipeline of signature extraction and curation, with no human assistance from the obtaining of company documents to the clustering of individual signatures. We use a sequence of heuristic methods, convolutional neural networks, generative adversarial networks and convolutional Siamese networks for signature extraction, filtering, cleaning and embedding respectively. We evaluate both the effectiveness of the pipeline at matching obscured same-author signature pairs and the effectiveness of the entire pipeline against a human baseline for document signature analysis, as well as presenting uses for such a pipeline in the field of real-world anti-money laundering investigation.
\end{abstract}

\section{Introduction}

The use of handwritten signatures for authentication and association is widespread throughout all industries, and particularly consistent in financial institutions. Recent developments have begun to digitise this process, enabling the storage of ‘online’ signatures which are captured along with dynamic information such as pen velocity, acceleration and pressure, as opposed to ‘offline’ signatures which are the conventional image left behind by the signer’s writing. The task of signature verification is to determine, given two captured signatures, whether both were signed by the same author. Online signatures are less challenging to verify due to the availability of more features, though previous research has shown success in verifying offline signatures \cite{DBLP:journals/corr/Dey0TGLP17}.

Signature verification is an important part of anti-money laundering investigations: the use of large numbers of ‘shell’ companies in order to obfuscate financial activities hinders investigation by generating large amounts of company documents that complicate the investigative process. When registering with bodies such as Companies House in the United Kingdom, companies are required to verify identities of signatories by providing signatures and can do so either electronically or by producing scanned signed documents. Illicit companies often choose the latter in order to limit the speed of investigations against them, as scanned documents are more challenging to extract information from automatically. Furthermore, a single signature is often used extensively and under different identities, suggesting that it may be used fraudulently enough times to render the ‘true’ identity meaningless. For that reason, we aim to solve this particular task by deciding whether a signature characterises an \emph{unskilled forgery} (a forgery made with no knowledge of the target signatory), rather than a \emph{skilled forgery} (a forgery intended to impersonate another's signature with prior knowledge of the signature form). An example of a typical document for signature analysis is given in Figure 1. 

Our main contributions can be summarised as follows:
\begin{itemize}
    \item We present an integrated pipeline combining deep learning and heuristic methods to perform the end-to-end process of signature investigation, previously reliant upon human extraction.
    \item We evaluate the resultant changes to the scalability, time and space cost of such investigations using the pipeline.
    \item We analyse the effects of including parts of the pipeline and of the application on different sources of signature data.
\end{itemize}

\begin{figure}
\begin{center}
\fbox{\includegraphics[width=\linewidth]{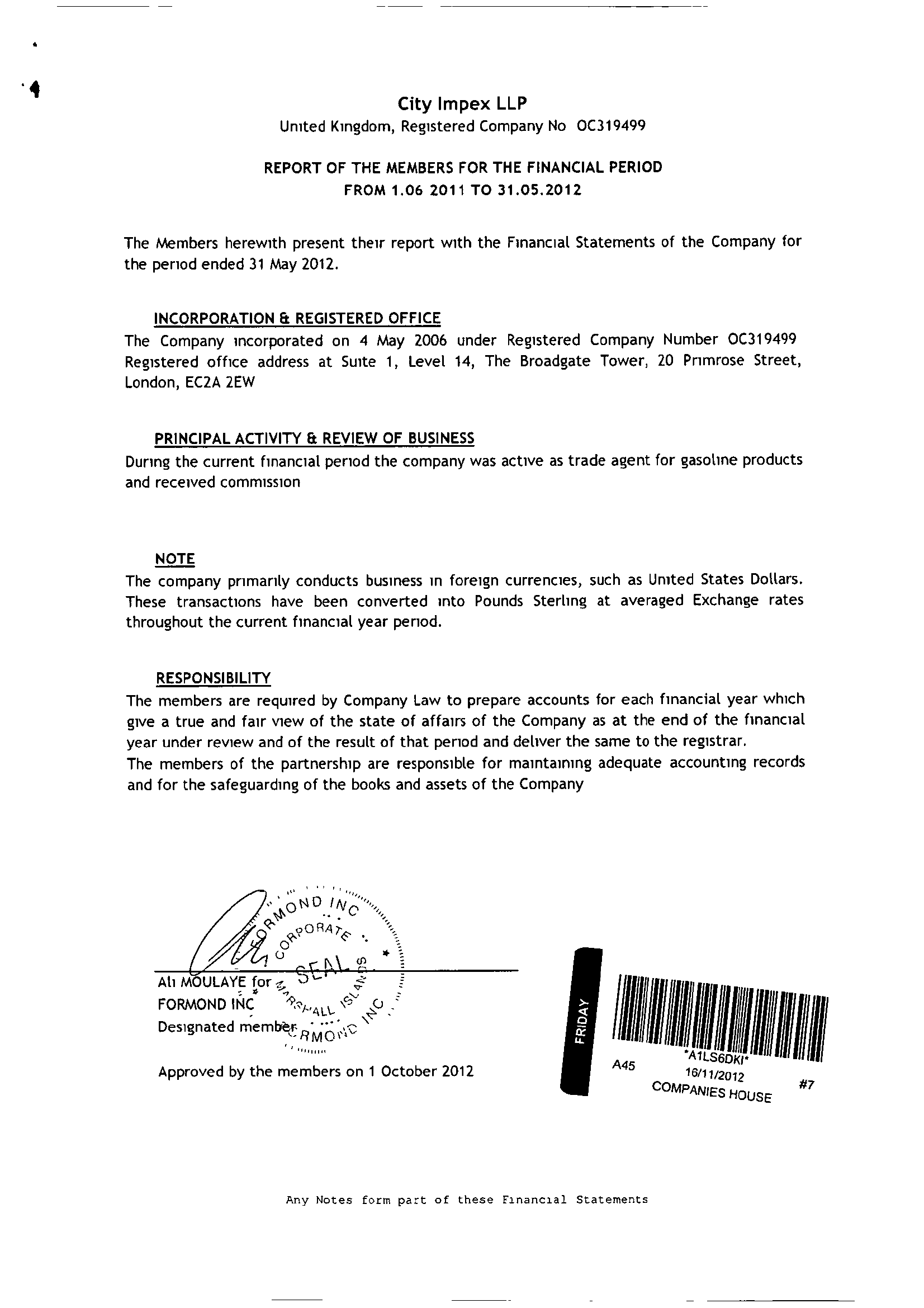}}
\end{center}
   \caption{A typical document containing a partially obscured signature.}
\label{fig:long}
\label{fig:onecol}
\end{figure}

The challenges inherent in signature analysis are primarily of time and accuracy. In the 2019-20 financial year more than 11 million documents were filed to Companies House, of which over 1.2 million were not electronically submitted and therefore likely to contain a handwritten signature\cite{companieshousedocs}. It is currently infeasible to extract useful information on signatures from this scale of documentation, due to the fact that the documents often have no usable metadata, enforced standards or other identification. In \cite{9151029} the human time cost of verifying a single signature against exactly one other took approximately 9 seconds per signature. In the context of AML, not only verification but identification is often required for analysis, which could lengthen the time cost. Furthermore, the time costs of obtaining the documents can be high, and the sum of these costs makes manual signature inspection a highly inefficient and unreliable process when targeted and an impossible task when not.

A significant part of manual signature investigation involves navigating inconsistent user interfaces, searching through documents of different formats for signature data that is often a small minority of the total content stored by the corporate data repository.

Accuracy also can vary significantly depending on the subject of the investigation. Figure 2 shows two examples of a known suspicious signature with significant variation. Any decision by a human or machine with respect to their match status brings implications for the sensitivity of the matching of other signatures: the more the model values recall (grouping similar signatures together) over precision (not grouping together dissimilar signatures), the smaller the number of possible distinct signatures in the image space becomes. If the number of distinct signatures on the image space is smaller than the number of actual signatories, then the complete-accuracy identification task may not be achievable. Human accuracy is often high in verifying against a single reference signature, but untested and unscalable to identification from a larger signature set. Furthermore, given that the signatures of interest are more likely than the general population to be individuals with an incentive to obfuscate or falsify signatures they write, the difficulty of recognising signatures will likely be higher.

\begin{figure}
\begin{center}
\includegraphics[width=\linewidth]{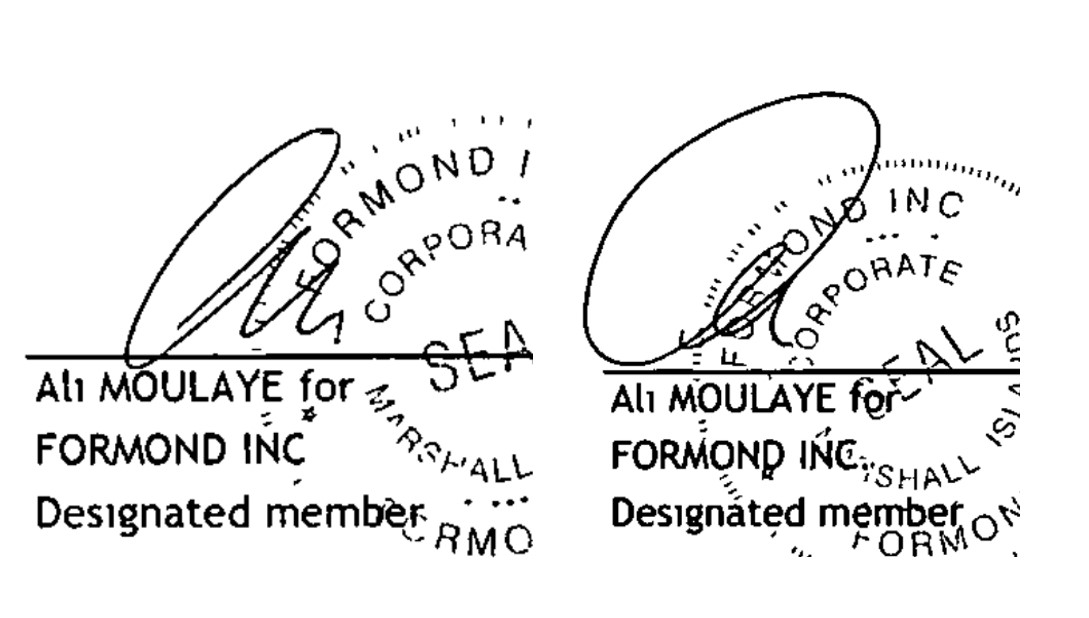}
\end{center}
   \caption{Two examples of a signature for the same company and year}
\label{fig:long}
\label{fig:onecol}
\end{figure}

A variety of datasets are used for reproducing the environment of the signature verification problem, including CEDAR, GPDS-960 and GPDS-4000. Each of these contains signatures and skilled forgeries from a wide range of authors; however, any real-world context is removed. As identified in \cite{inproceedings}, most studies do not consider the extraction stage: the signatures are signed isolated on standardised spaces, unlike signatures on financial declarations which are frequently over document text, form elements or under imprinted stamps. The discrepancy between the signature dataset used for academic purposes and the more common context of real-world signatures restricts the usefulness of signature verification models, as strong performance can only be reliably obtained on signatures which have been curated explicitly for the model. Given the vast signature data that has been generated before the existence of signature verification algorithms, it would be beneficial to develop systems that are resilient to the challenges of unrefined signatures in context.

Previous research has explored the effectiveness of signature verification methods on document-obtained signatures \cite{9151029}, but this process requires manually collected signature images, detected and cropped from real-world documents, limiting the speed and scope of the overall process.

We use an integrated pipeline of methods to perform the processes of signature extraction, verification and clustering, from a set of document images to a set of signature clusters, where a cluster represents a group of signature images with a common signatory. We use connected component analysis, supervised by a convolutional neural network to extract signature images from the document set. The outputs of this stage, the set of signature images extracted, are then refined using a CycleGAN to remove real-world context such as text, form elements and stamps. From the resultant signature images, feature vectors are obtained using a convolutional Siamese neural network, which are then clustered using hierarchical clustering to obtain signatory information.

\section{Related Work}

\subsection{Signature De-noising}

In order to make signatures readable for the algorithm, they must be de-noised first to remove any potentially unnecessary components that were extracted. One of the common methods used to obfuscate signatures is to stamp over the signature (Figure 1). A CycleGAN \cite{DBLP:journals/corr/ZhuPIE17} \cite{9151029} was used to remove stamps from extracted signature images. The model was trained on a set of unpaired examples of signatures labelled as ‘clean’ (does not contain a stamp) or ‘unclean’ (contains a stamp). The model achieved strong results on the financial dataset which was used, and of which 26.3\% were clean (in our application on Companies House, less than 1\% contain no obstructing features). The dataset used contained two types of document, \emph{order documents} and \emph{signature declaration documents}, whereas our problem involves extracting signatures from any corporate document on Companies House. Furthermore, the \emph{signature declaration documents} contain signatures which are unstamped and repeated three times, used by the authors as reference signatures. Our Companies House data source does not provide reference signatures, and therefore each signature must be treated equally.

\subsection{Signature Similarities}

In \cite{DBLP:journals/corr/Dey0TGLP17} the authors use a neural network that applies convolutional operations on two images independently and a comparison operation to compute similarity between two signatures. They obtain high (100\%) accuracy on the CEDAR dataset and 78\% to 87\% accuracy when trained and evaluated on other datasets. When trained on one dataset and evaluated on another, the accuracy is often much lower and less consistent (53\% to 95\%), suggesting high variation in the compositions of the standard signature datasets.

\section{Proposed Method}

Our method for signature analysis derives from a set of document images, a set of signature clusters. This uses an integrated pipeline of several stages including extraction, cleaning, feature extraction and clustering. Figures 3-5 show the interactions between the major stages in the pipeline.
\begin{figure}[t]
\begin{center}
\includegraphics[width=\linewidth]{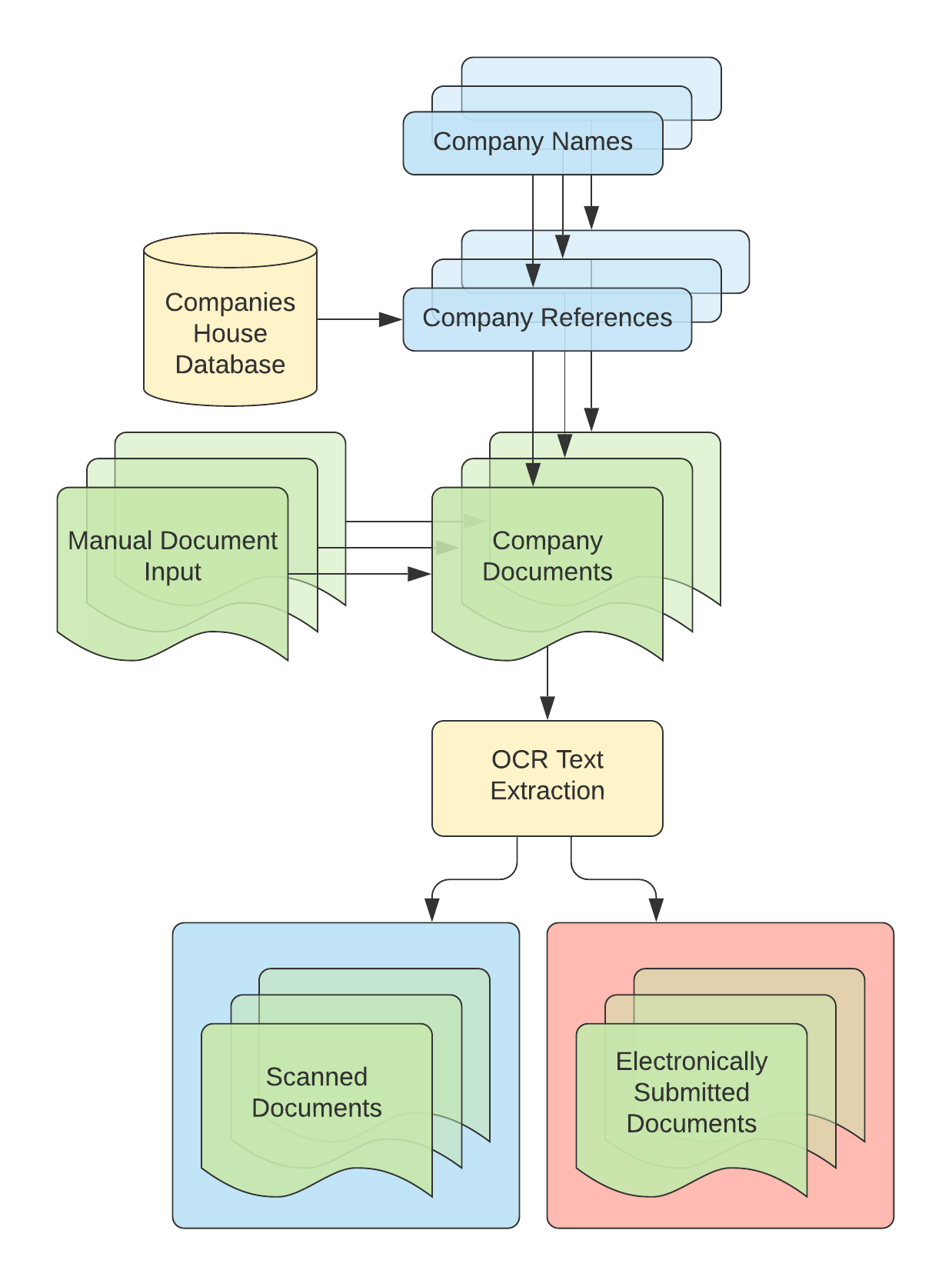}
\end{center}
   \caption{The document retrieval stage of the pipeline}
\label{fig:long}
\label{fig:onecol}
\end{figure}
\begin{figure}[t]
\begin{center}
\includegraphics[width=0.5\linewidth]{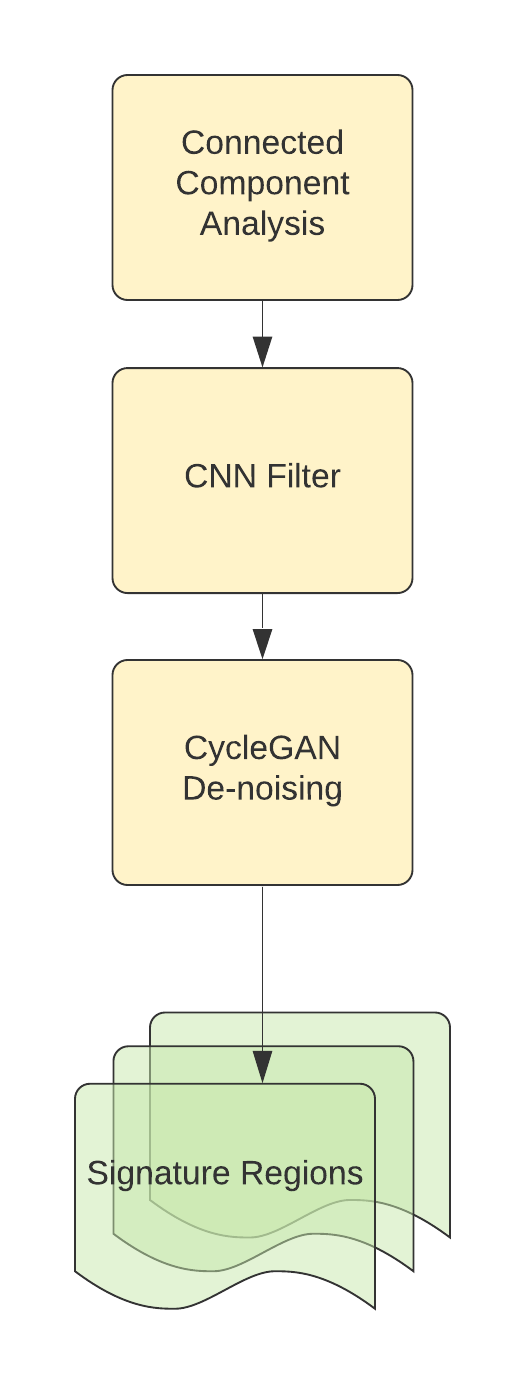}
\end{center}
   \caption{The enhancement (extraction, filtering, denoising) stages of the pipeline}
\label{fig:long}
\label{fig:onecol}
\end{figure}
\begin{figure}[t]
\begin{center}
\includegraphics[width=\linewidth]{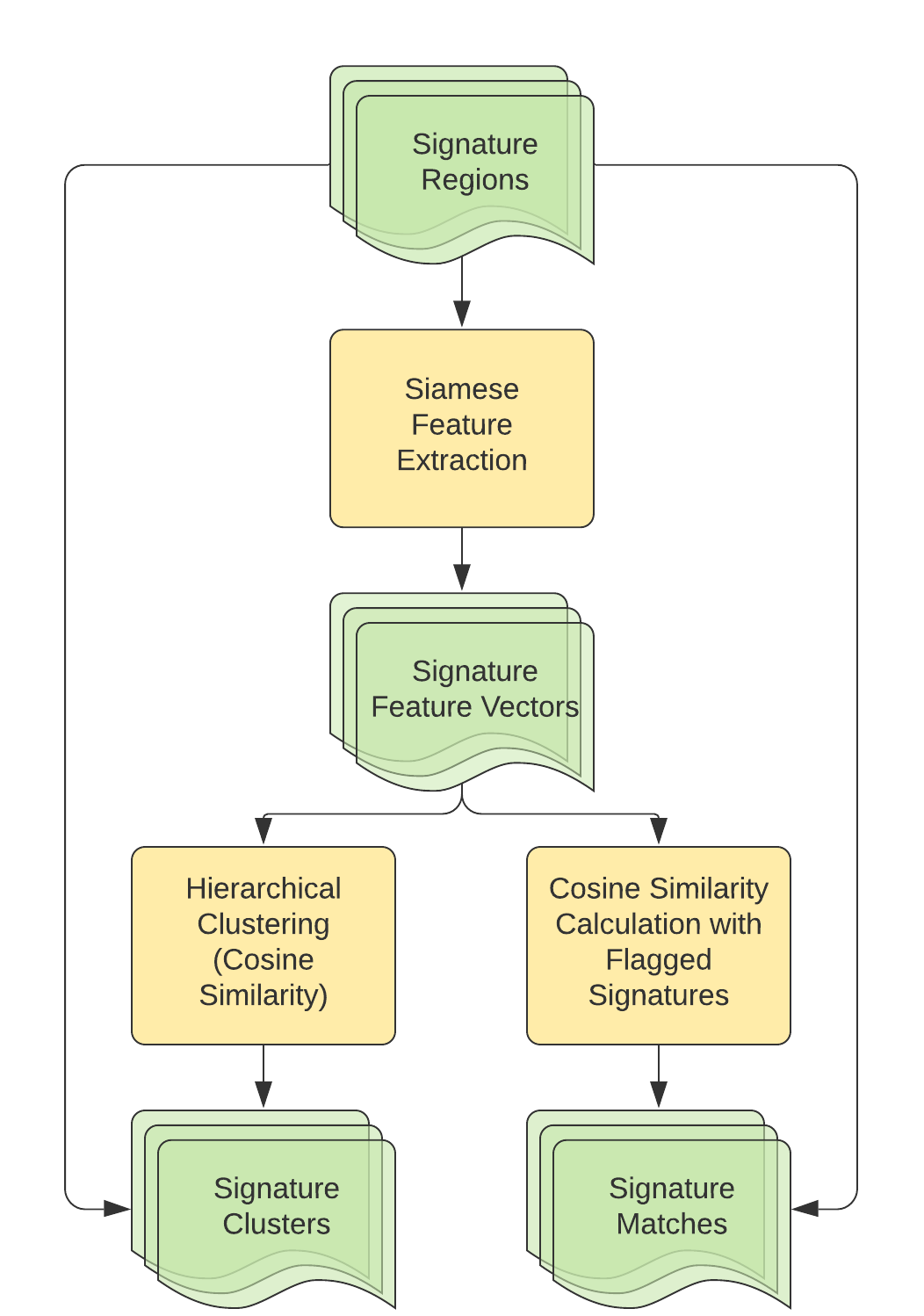}
\end{center}
   \caption{The identification (feature extraction, clustering, matching) stages of the pipeline}
\label{fig:long}
\label{fig:onecol}
\end{figure}

\subsection{Optical Character Recognition}

The pipeline extracts signatures from each page in a set of company documents, and therefore must handle the case of no signatures existing in an image. In order to reduce the false positive rate for extraction of signatures, we use the Tesseract optical character recognition engine \cite{10.5555/1288165.1288167} to scan for key words and phrases that indicate no signature is present (for example, ``electronically filed" is present on most electronically filed documents on Companies House). This prevents the extraction stage from identifying signatures where none exist in the file, but does not prevent false positive signature candidate regions where genuine signatures exist in the file.

\subsection{Connected Component Analysis}

Connected component analysis uses a graph representation of an image to extract high-level information and separate regions which contain similar characteristics. A connected region is defined as a set of pixels in which no adjacent pixel differs significantly from its neighbour. Signatures typically are sets of connected regions in close proximity. We first decompose the binarized document image into set of connected regions, then merge nearby regions while applying heuristic filters on attributes such as density and aspect ratio, and apply edge removal in order to remove common obstructions. We merge connected regions because many signatures are not entirely connected and consist of several large connected regions; after merging the bounding boxes of the regions are obtained by finding the minimum and maximum \(x\)  and \(y\) locations of the connected vertices.

The heuristic methods and constraints used operate only under the assumption that the image is a scan of an A4-printed and primarily text- and form-occupied document, the same format as all documents found on Companies House and most corporate registers.

\subsection{CNN Filtering}

Connected component analysis is an efficient and effective method of extracting signature images, but it is often imprecise, returning also regions which are not signatures: for example, addresses, dates or other handwritten sections. The nature of connected component analysis means that it is unrealistic to expect the method to be able to distinguish between these without highly precise formulations of the characteristics of handwritten text and signatures, which is robust to changes in authorship and writing style. Therefore, we train a convolutional neural network to predict whether a given image is a signature or not. Each output from the CCA stage is passed through the CNN, and retained if the result (using sigmoid activation) is over a given threshold, else it does not progress to the next stage of the pipeline.

The neural network receives in the input layer an binarized image vector of shape \((256, 256)\) and range \([0, 1]\), and through a series of convolutional, pooling and fully-connected layers generates an output vector of shape \(1\) in the range \([0, 1]\).

\subsection{CycleGAN Cleaning}

In \cite{9151029} the authors use a CycleGAN to remove stamps from signature images so that they may be verified accurately. We use the CycleGAN in a similar way to remove all non-signature data (including text, form elements and stamps), by training it on a combination of manually cleaned (paired) and unpaired examples in order to optimise training. The CycleGAN architecture is standard (taking images of shape \(256 \times 256\), except for an alteration to take as input and return as output greyscale images. In this alteration, the generator produces an output of shape \((256, 256, 1)\) and the discriminator takes this shape as input, and produces a singular output; intuitively, the likelihood that the image was generated by the generator.

The CycleGAN uses the cycle consistency loss function. Given two domains $X$ and $Y$ (in this application, $X$ is the set of raw signature images and $Y$ the set of isolated signature images), we aim to generate two mapping functions $F$ and $G$ such that $F:X\longrightarrow Y$, $F(G(X)) \approx X$ and vice versa. This is achieved by minimising the loss function:
\begin{equation}
\begin{split}
\mathcal{L}_{cyc}(G, F) = \mathbb{E}_{x \sim p_{data}(x)}||F(G(x)) - x||_{1}\\ + \mathbb{E}_{y \sim p_{data}(y)}||G(F(y)) - y||_{1}
\end{split}
\end{equation}

It was decided to use a combination of manually cleaned and unpaired examples, despite the CycleGAN's ability to handle unpaired examples, primarily due to the problem of mode collapse in which one of the adversarial components of the CycleGAN (either the discriminator or the generator network) dominates the other in terms of accuracy, causing the gradient of its output to tend to zero. Including paired examples in unpaired training environments has also been found to increase the quality of the domain mapping function\cite{GINGER202050}.

\subsection{Feature Extraction}

In \cite{DBLP:journals/corr/Dey0TGLP17} and \cite{9151029}, the authors use a convolutional Siamese neural network to compute the probability of a common signatory in two signature images. We use the VGG16\cite{Simonyan2015VeryDC} model, pre-trained on ImageNet, for the encoder stage and the second-to-last fully-connected layer outputs as the embedding of shape \((4096)\), on account of its high accuracy in \cite{9151029}. The model is trained on a manually collected dataset of automatically extracted and cleaned signatures from Companies House, in paired examples \(((X_1, X_2), y)\), where \(X_1, X_2\) are two cleaned signature images and \(y \in \{0, 1\}\) the author similarity label.

The Siamese neural network is constructed first by initialising the VGG16 network with ImageNet weights, and removing all layers beyond the second-to last fully-connected layer: we refer to this as the encoder network. The complete Siamese network is then constructed which receives as input a pair of images: a vector of shape \((2, 224, 224)\), repeated along each RGB axis, propagates each image vector independently through the encoder network and concatenates the results to obtain a vector of shape \((2, 4096)\), and subsequently through a cosine similarity layer to return a vector of shape \((1)\) (no activation function is applied) as shown in Figure 7.

In order to train the encoder network, we first train the complete Siamese neural network on the generated dataset of paired image examples and the corresponding author similarity labels, using the binary cross-entropy loss function. We then construct a new encoder network initialised to the weights of the encoder layers in the complete Siamese network in order to generate embeddings from single images, whose match can be identified using the post-concatenation stages of the network.

\begin{figure*}
\begin{center}
\includegraphics[width=\linewidth]{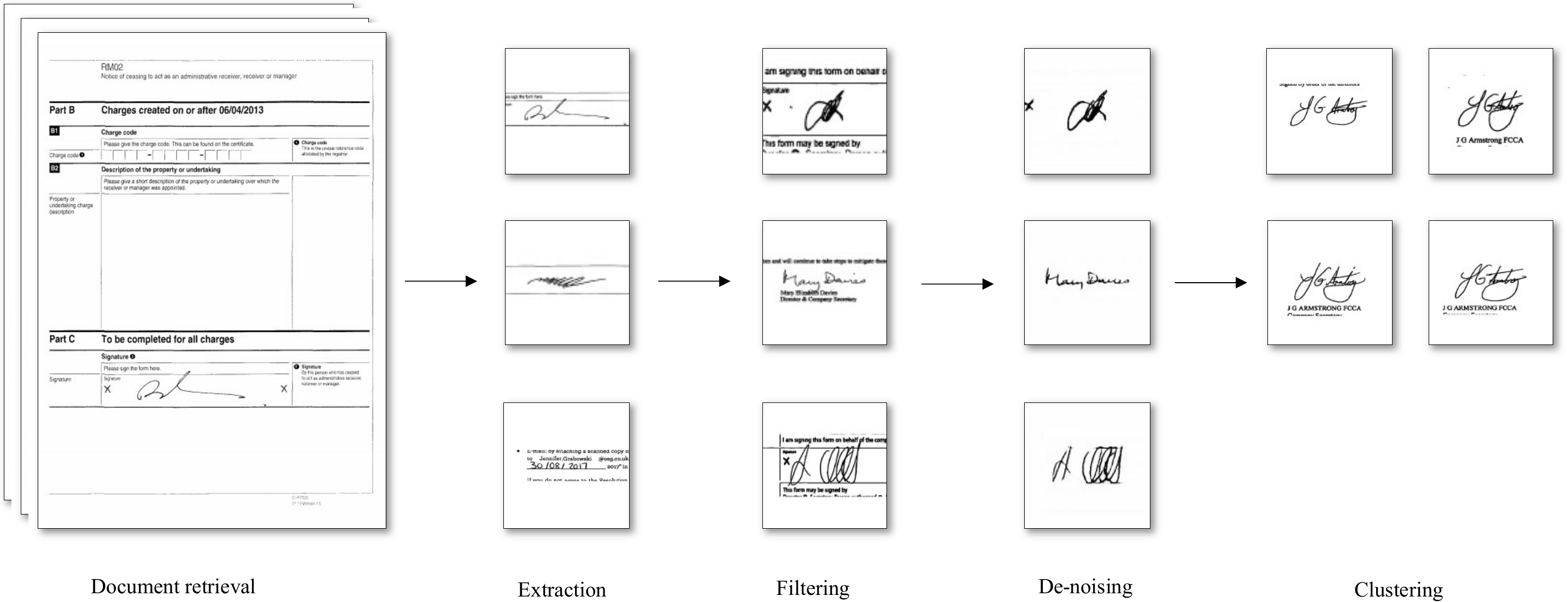}
\end{center}
   \caption{Examples of the staged outputs}
\label{fig:long}
\label{fig:onecol}
\end{figure*}

\subsection{Hierarchical Clustering}

The Siamese neural network is trained to generate feature vectors that have high cosine similarity where signatures are by the same author, and low cosine similarity elsewhere. Therefore, we can apply hierarchical clustering using cosine similarity as the distance metric in order to obtain a set of signature clusters which contain signatures by the same author, and therefore are verified.

The hierarchical clustering algorithm uses a threshold $t$, which is imposed upon the clusters as a criterion that no two vectors in the same cluster can have a distance metric greater than $t$. Given that the clustering algorithm uses cosine similarity as the distance metric, and that the Siamese neural network was trained to produce vectors in which the cosine similarity matched the semantic author similarity, then it is reasonable to set $t$ as the required human confidence level for matching any two signatures.

\section{Training}
The filtering, cleaning and feature extraction stages rely on the use of trained machine learning models. The models require different specifications of training datasets.

\subsection{CNN Filtering}

The convolutional neural network validating the CCA output required a set of labelled images of signature regions and non-signature regions. A custom dataset was generated with the manually-labelled outputs of the CCA stage, combined with extracted signatures from the Tobacco-800 dataset. The dataset was augmented while training using the library Keras\cite{chollet2015keras}, with random cropping and rotation applied to images after being drawn from the dataset and before training. Augmentation was not applied for the validation experiments. The model was trained for 3000 epochs and achieved a final validation accuracy of 93.5\%.

\subsection{CycleGAN Cleaning}

The CycleGAN used to remove obstructive features from the signature regions required a set of labelled images of clean and unclean signature regions. A custom dataset was generated using a combination of manually cleaned examples, manually labelled Tobacco-800 signatures, and manually-labelled extracted signatures from Companies House to equalise the datasets. The model was trained for 200 epochs.

\subsection{Siamese Feature Extraction}

The Siamese neural network required a set of image pairs labelled with their binary author similarity. This dataset was generated from a Companies House signature dataset in which all images are labelled with their author, after applying the CycleGAN cleaning. The model was trained for 850 epochs and achieved a validation accuracy of 81.4\%. The validation dataset and the training dataset did not share authorship.

\begin{figure*}
\begin{center}
\includegraphics[width=0.4\linewidth]{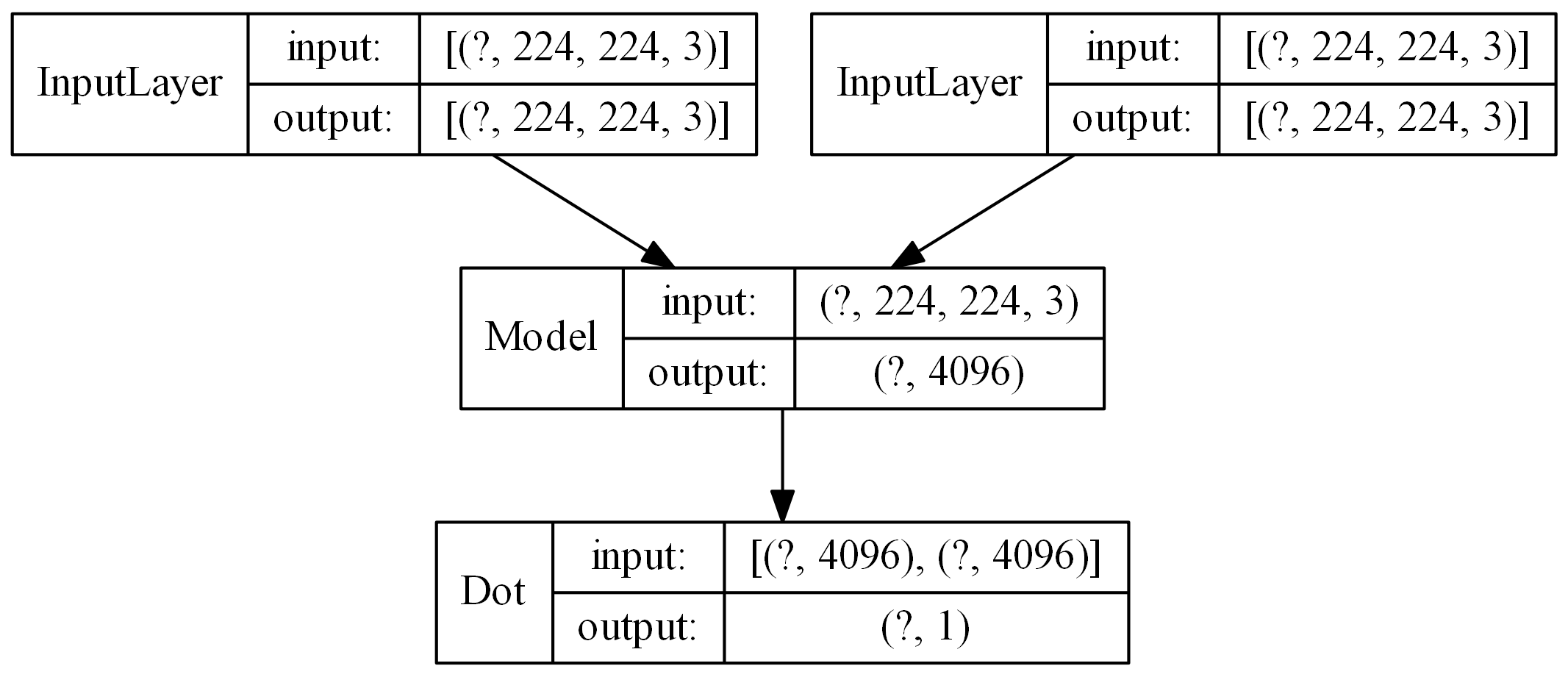}
\end{center}
   \caption{The format of the Siamese Neural Network}
\label{fig:long}
\label{fig:onecol}
\end{figure*}

\section{Results}

\subsection{Examples}

Examples of the output images at each major stage are shown in Figure 6.

\subsection{Signature Verification}

For the verification task, we take the author signatures from each of the CEDAR, Tobacco-800 and randomly sampled Companies House signatures and generate one-to-one comparisons, in which the Siamese model predicts a match probability. We find that the model performs best on the CEDAR dataset, which contains no obstructive features other than the signature, and that applying the cleaning process reduces the accuracy of the Siamese neural network where the signature is already clean. However, this analysis only shows the average effects on a large and diverse sample of signatures. While the cleaning process degrades the accuracy of most signatures, the absence of a cleaning process renders the identification of particular signatures routinely obscured virtually impossible; in addition, given the nature of the target usage for the pipeline of anti-money laundering, it is likely that the minority of signatures which are obstructed by stamps are more likely to be of interest, and therefore the value of the pipeline depends upon maintaining a level of accuracy consistent on inputs of low quality, rather than a maximised expected accuracy over all groups combined. The receiver operating characteristic curves for the inference on each signature dataset are shown in Figure 8.

\begin{figure}
\begin{center}
\includegraphics[width=\linewidth]{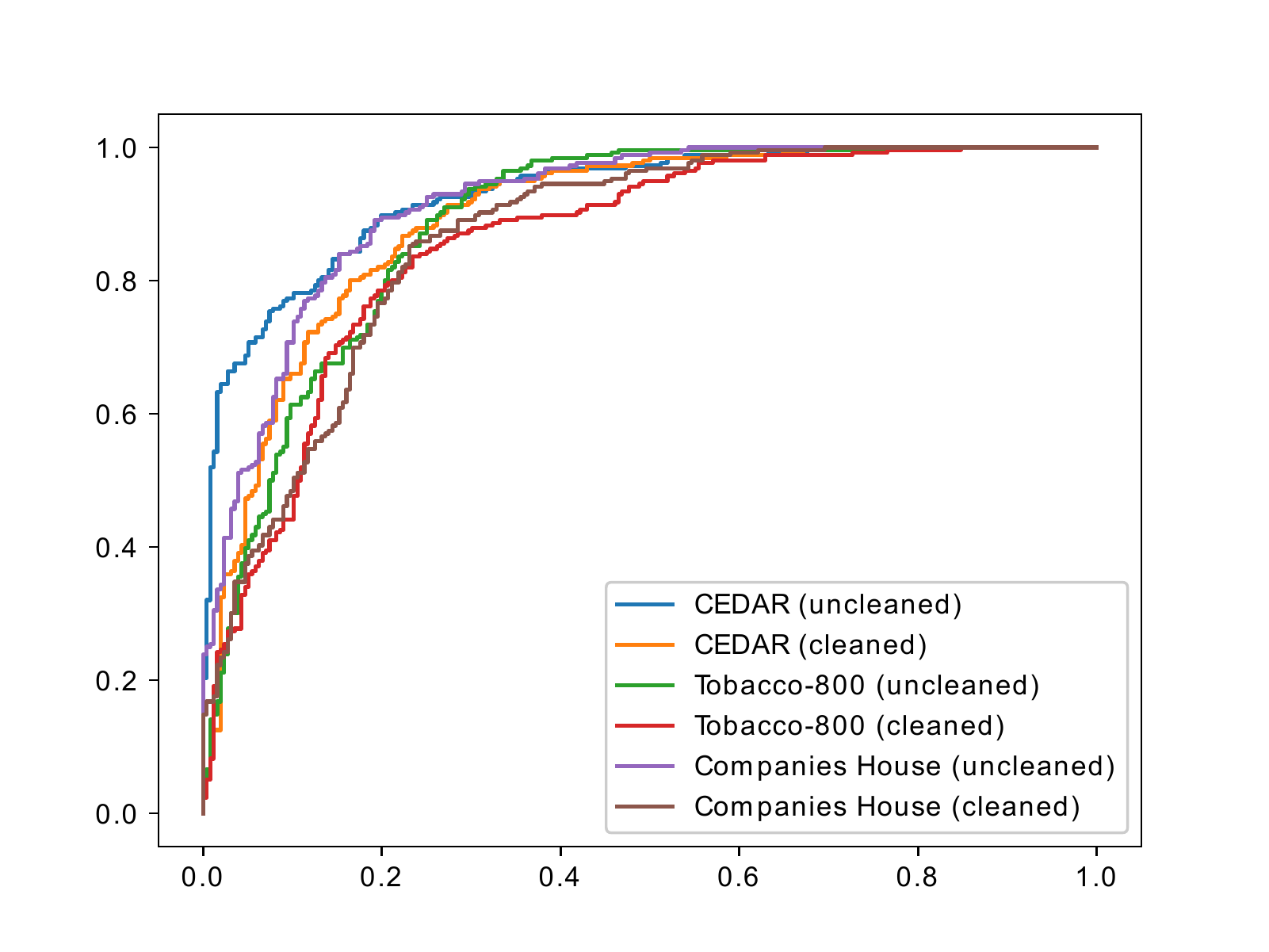}
\end{center}
   \caption{Receiver operating characteristic curves by type and origin of signature}
\label{fig:long}
\label{fig:onecol}
\end{figure}

\subsection{Optical Character Recognition}

We find that using optical character recognition to filter out documents slightly increases the precision of the signature extraction phase, from a manual evaluation of 266 signatures extracted from a random sample of companies, shown in Table 1.

\begin{table}
\centering
 \begin{tabular}{||c c c||} 
 \hline
 Metric & OCR & No OCR \\ [0.5ex] 
 \hline
 Extraction precision & 0.947 & 0.935 \\
 \hline
\end{tabular}
\caption{Precision metrics with and without OCR}
\end{table}

\subsection{Signature Data Collection Scalability}

The challenges of scalability are primarily of time and space. Time is an finite and expensive (human time more so) used in signature investigations. In \cite{coetzer:inria-00103737} an experiment on human signature verification found that on a large dataset ($n = 765$) of signatures a human took on average between 3.5s and 4.7s to verify each signature against a reference signature, when instructed not to spend extensive amounts of time on any signature. On a similar dataset ($n = 537$), the fully automated pipeline took on average 0.1s per signature in the clustering of signatures, suggesting a substantial performance improvement. Furthermore, computer time is both less expensive and more easily available, and would likely be preferable to human time on this task even with all else equal.

On the challenge of space, from a sample of 266 signatures collected from randomly sampled companies from Companies House, the storage of each signature image under JPEG compression resulted in an average file size of 26.8KB, compared to the average signature embedding file size of 3.0KB: a reduction of 89\%, suggesting the efficiency gains of using the reduced representation for collecting signature data.

\subsection{Clustering Accuracy}

We evaluate the accuracy of the overall clustering process by manually partitioning the extracted signatures into distinct author clusters and computing the Adjusted Rand Index between the human-defined and machine-predicted clusters. The Adjusted Rand Index is a modification of the Rand Index, which is a measure of the similarity between two clusterings, calculated as the proportion of possible pairings in each clustering in which the co-occurrence of both elements within a cluster is identical in both clusterings:
\begin{equation}
    \frac{TP + TN}{TP + TN + FP + FN}
\end{equation}
where $TP, TN, FP, FN$ are the numbers of true positives, true negatives, false positives and false negatives respectively. The Adjusted Rand Index adjusts to consider the permutations of cluster names, and as found by \cite{ari} is a sufficient measure of accuracy of clusterings. We can therefore compare the accuracy of the signature clustering process from the signatures of a randomly selected company against the accuracy findings in \cite{9151029} which used manually-extracted input images of a similar quality. We find a slight improvement in accuracy over the baseline (using the VGG16 model), as shown in Table 2.

\begin{table}
\centering
 \begin{tabular}{||c c||} 
 \hline
 Method & Accuracy \\ [0.5ex] 
 \hline
 Human Majority Voting\cite{9151029} & 91.66 \\
 Human Individual Voting\cite{9151029} & 89.25 \\
 Fully automated pipeline & 78.19 \\
 Engin, Kantarci et al\cite{9151029} & 76.38 \\
 \hline
\end{tabular}
\caption{Class pairing accuracies}
\end{table}

Note that although the accuracies are broadly comparable, they have been obtain through tests on different materials. Our pipeline is applied to images of a lower quality, with more frequent obfuscation, and on documents with higher amounts of non-signature handwritten content. The correct clusters were determined using the names found near signatures where necessary.

\section{Conclusion}

Signatures used as authentication for corporate documents on Companies House have high intra-class variation, high inter-class similarity and high rates of intersection with non-signature image data. Previous research has shown the application of deep learning models and heuristic methods to be effective in combination with human-performed actions at extracting high-level information from offline signatures, with effectiveness decreasing as a greater proportion of the human workload is automated. We find that a combination of heuristic and deep learning models can approximate the complete workload of extracting high-level authorship information from corporate filing data with medium to high accuracy, and significantly increased speed and scalability when compared to a human baseline.

\subsection{Domain Extension}

We also find that both existing and novel methods of handwritten signature processing extend well to larger and less supervised problem domains with increased diversity and decreased data quality, such as signature clustering from corporate documents. Expansions of the problem domain to these less constrained tasks increase the range of possible applications of existing methods and equip human investigators with more relevant and usable high-level information from the large, low-level input datasets, for example by enabling anti-money laundering investigators to more efficiently explore the signature representation space of a population of documents.

\subsection{Applications}

The pipeline described in this research has several uses in the real-world field of anti-money laundering investigation. Primarily, as an exploratory tool for extracting signature clusters from document collections, the results of which can be used by investigators both to more quickly identify known suspicious signatories such as the example given of Ali Moulaye, or extract the main signatories of interest without needing to manually search through the document set. The secondary use of the pipeline is in conducting a search of a document set for a given set of signatures. This enables investigators with known target signatories to save time and labour by avoiding manually searching the documents and other signatories for the presence of a given signature.

Despite the accuracy of the pipeline being lower than that of a human investigator, the pipeline retains its usability as an aide in evidence-gathering. Existing signature investigations are largely driven by manual inspection, which is efficient at establishing key signatures over time spent searching documents, but less so at gathering large datasets of signature examples to prove anomalousness in signature activity. This pipeline is able to carry out the task more efficiently than such manual investigations.

Furthermore, the embedding extraction stage presents a number of potentially useful applications not already used in investigations. Properties of the distribution of the embedding set extracted, for example the degree of concentration of signature embeddings, could be used in investigation of the documents as a collection, rather than the signatures on individual documents.

The pipeline developed in this research was targeted exclusively at Companies House in the United Kingdom, but it is trivial to extend to other corporate data repositories. However, if the distribution of signature data is significantly different in other data sources, this may require either retraining of the models or additional data processing steps.

\subsection{Further Research}

Further research into this area is needed to determine the limits of the resilience to systemic low quality of input data from Companies House provided by deep learning techniques. A significant unknown factor in need of further research is the theoretical limit to the accuracy of large-scale signature identification, given the high intra-class variation and inter-class similarity inherent in the signatures of large populations, as well as any possible strategies to overcome or mitigate this challenge. In addition, the methods proposed in this research extracted text from each document for the purposes of accuracy improvement only, in order to vet signature images by the contents of their accompanying text: further investigation could reveal whether the application of natural language processing techniques on the extracted text could provide additional information on the identities and relationships between extracted signatures.

Furthermore, the nature of the domain that a signature verification model is applied to may significantly affect the accuracy and coverage requirements. Anti-money laundering is an adversarial domain, in which the subjects providing the signature are incentivised to minimise the accuracy and speed of recognition. This is different to usual, crime-free settings such as in legitimate business, where the subjects have no such negative incentive. Therefore future research might identify whether the application of specific adversarial learning techniques might improve the resilience of the pipeline to characteristics commonly found in the signatures of persons of interest.

{\small
\bibliographystyle{iccv_21_template/ieee_fullname}
\bibliography{egbib}
}

\end{document}